\documentclass[final,5p,times]{elsarticle}

\usepackage{graphicx}
\usepackage{epsfig}

\usepackage[ruled,vlined]{algorithm2e}

\usepackage{amssymb}
\usepackage{amsmath}

\usepackage{hyperref}

\usepackage{subfigure}
\usepackage{dblfloatfix} 
\usepackage{multirow} 
\usepackage{booktabs}

\usepackage{balance}

\newcommand{\hide}[1]{}

\newcommand{\dquote}[1]{``#1''}
\newcommand{\etal}{\textit{et al.}~}

\usepackage{colortbl}
\definecolor{red}{rgb}{0,0,0}

\journal{Computer Vision and Image Understanding}

\begin{document}

\begin{frontmatter}


\author[micc,stanford]{Lamberto Ballan\corref{cor1}}
\ead{lamberto.ballan@unifi.it}
\author[micc]{Marco Bertini}
\ead{marco.bertini@unifi.it}
\author[unimore]{Giuseppe Serra}
\ead{giuseppe.serra@unimore.it}
\author[micc]{Alberto Del Bimbo}
\ead{alberto.delbimbo@unifi.it}

\cortext[cor1]{Corresponding author. Tel.: +39 055 2751395.}
\address[micc]{Media Integration and Communication Center (MICC), Universit\`a degli Studi di Firenze, Viale Morgagni 65, 50134 Firenze, Italy}
\address[unimore]{Dipartimento di Ingegneria \dquote{Enzo Ferrari}, Universit\`a degli Studi di Modena e Reggio Emilia, Via Vignolese 905/b, 41125 Modena, Italy}
\address[stanford]{Computer Science Department, Stanford University, 353 Serra Mall, Stanford, CA 94305, United States}


\title{A data-driven approach for tag refinement and localization in web videos}

\begin{abstract}
Tagging of visual content is becoming more and more widespread as web-based services and social networks have popularized tagging functionalities among their users. These user-generated tags are used to ease browsing and exploration of media collections, e.g.~using tag clouds, or to retrieve multimedia content. However, not all media are equally tagged by users. Using the current systems is easy to tag a single photo, and even tagging a part of a photo, like a face, has become common in sites like Flickr and Facebook. On the other hand, tagging a video sequence is more complicated and time consuming, so that users just tag the overall content of a video. 
In this paper we present a method for automatic video annotation that increases the number of tags originally provided by users, and localizes them temporally, associating tags to keyframes.
Our approach exploits collective knowledge embedded in user-generated tags and web sources, and visual similarity of keyframes and images uploaded to social sites like YouTube and Flickr, as well as web sources like Google and Bing. Given a keyframe, our method is able to select ``on the fly'' from these visual sources the training exemplars that should be the most relevant for this test sample, and proceeds to transfer labels across similar images. Compared to existing video tagging approaches that require training classifiers for each tag, our system has few parameters, is easy to implement and can deal with an open vocabulary scenario. We demonstrate the approach on tag refinement and localization on DUT-WEBV, a large dataset of web videos, and show state-of-the-art results.
\end{abstract}


\end{frontmatter}


\section{Introduction}\label{sec:intro}

Over the past recent years social media repositories such as Flickr and YouTube have become more and more popular, allowing users to upload, share and tag visual content.
Tags provide contextual and semantic information which can be used to organize and facilitate media content search and access.
The performance of current social image and video retrieval systems depends mainly on the availability and quality of tags. However, these are often imprecise, ambiguous and overly personalized \cite{kennedy-2006}. Tags are also very few (typically three tags per image, on average) \cite{zwol-2008}, and their use may change over time, following the creation of new folksonomies created by users. Another issue to be considered is the `web-scale' of data, that calls for efficient and scalable annotation methods.

\textcolor{red}{Many efforts have been done in the past few years in the area of content-based tag processing for social images \cite{csur-2012,ssurvey}.
The main focus of these works has been put on three aspects: \emph{tag relevance} (or \emph{ranking}) \cite{liu-2009}, \emph{tag refinement} (or \emph{completion}) \cite{dliu-2010} and \emph{tag-to-region localization} \cite{mori-2010}.
Among the others, nearest-neighbor based approaches have attracted much attention for image annotation \cite{makadia-2008,guillaumin-2009,2pknn-2012,ballan-2014}, tag relevance estimation \cite{li-2009} and tag refinement \cite{mtap-2014}. Here the key idea is that if different users label similar images with the same tags, these tags truly represent the actual visual content. So a simple voting procedure may be able to transfer annotations between similar images.}
This tag propagation can be seen as a lazy local learning method in which the generalization beyond the training data is deferred until test time. A nice property of this solution is that it naturally adapts to an open vocabulary scenario in which users may continuously add new labels to annotate the media content. In fact, a key limitation of the traditional methods in which classifiers are trained to label images with the concept represented within, is that the number of labels must be fixed in advance.
More recently, some efforts have been made also to design methods to automatically assign the annotated labels at image level to those derived semantic regions \cite{mori-2010,yang-2011,sebe-2013}. A relevant example is the work of Yang \etal \cite{yang-2011} in which the encoding ability of group sparse coding is reinforced with spatial correlations among regions.

The problem of \emph{video tagging} so far has received less attention from the research community. Moreover, typically it has been considered the task of assigning tags to whole videos, rather than that of associating tags to single relevant keyframes or shots.
Most of the recent works on web videos have addressed problems like: \emph{i) near duplicate detection}, applied to IPR protection \cite{douze-2010,jingkuan-2013} or to analyze the popularity of social videos \cite{liu-2013}; \emph{ii) video categorization}, e.g.~addressing actions and events \cite{sclaroff-2009,snoek-2014}, genres \cite{wu-2012} or YouTube categories \cite{ytcat-2010}.
However, the problem of video tagging ``in the wild'' remains open and it might have a great impact in many modern web applications.

\begin{figure}[!t]
\centering
\includegraphics[width=1\columnwidth]{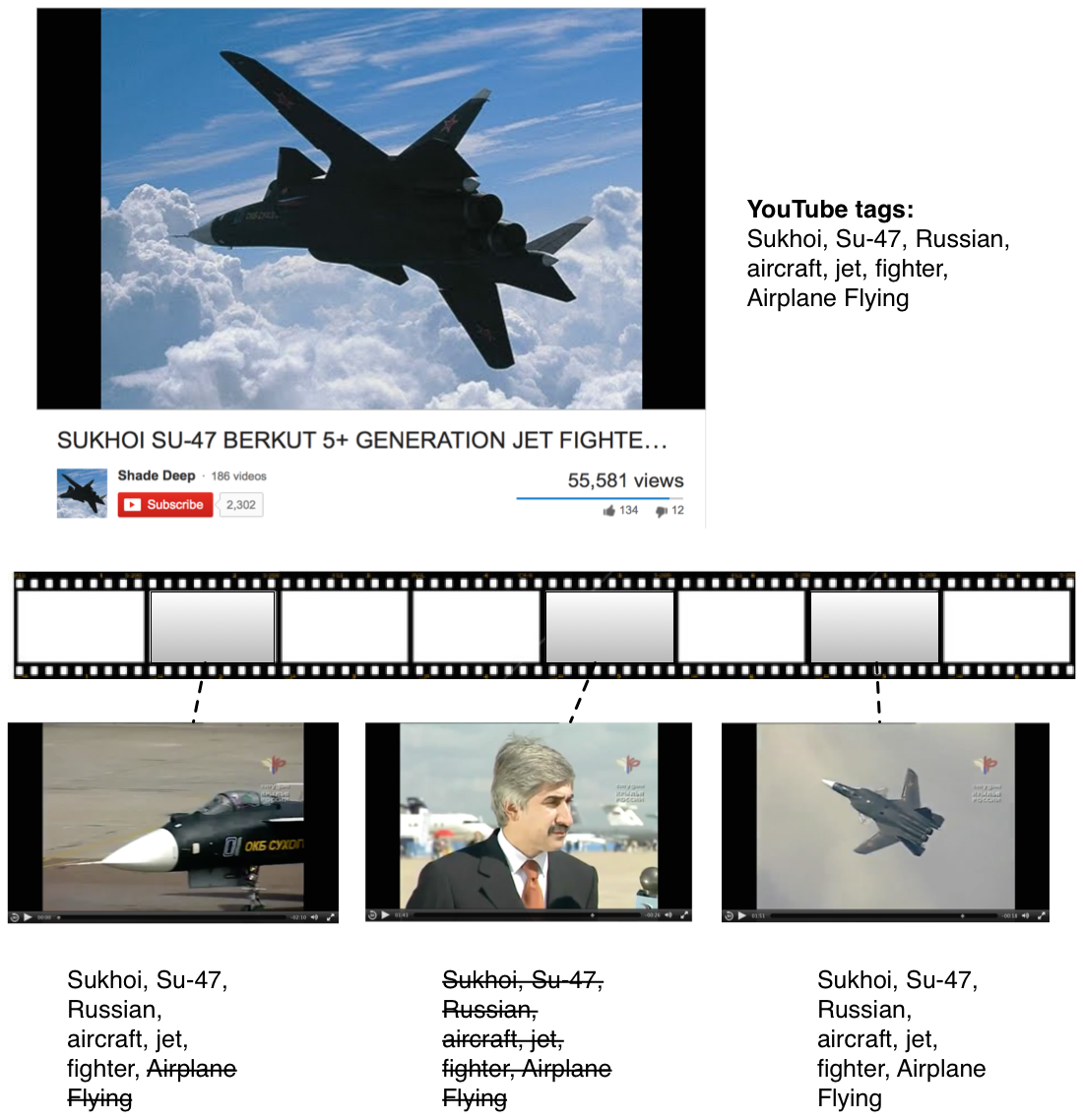}
\caption{Example of video tag localization: \emph{top)} YouTube video with its related tags; \emph{bottom)} localization of tags in keyframes.}
\label{fig:exampleYoutube}
\vspace{10pt}
\end{figure}

\textcolor{red}{In this paper, the proposed method aims at two goals: to extend and refine the video tags and, at the same time, associate the tags to the relevant keyframes that compose the video, as shown in Fig.~\ref{fig:exampleYoutube}.
The first goal is related to the fact that the videos available on media sharing sites, like YouTube, have relatively few noisy tags that do not allow to annotate thoroughly the content of the whole video. Tackling this task can be viewed also as an application of image tag refinement to video keyframes \cite{ssurvey,dliu-2010}.}
The second goal is related to the fact that tags describe the global content of a video, but they may be associated only to certain shots and not to others. Our approach takes inspiration from the recent success of nonparametric data-driven approaches \cite{makadia-2008,hays-2008,vicente-2011,liu-2011}. 
We build on the idea of nearest-neighbor voting for tag propagation, and we introduce a temporal smoothing strategy which exploits the continuity of a video. Compared to existing video tagging approaches in which classifiers are trained for each tag, our system has few parameters and does not require a fixed vocabulary. Although the basic idea has been previously used for image annotation, this is the first attempt to extend this idea to video annotation and tag localization.
\smallskip

Our contributions can be summarized as follows:
\begin{itemize}
\item We propose an automatic approach that locates the temporal positions of tags in videos at keyframe level. Our method is based on a lazy learning algorithm which is able to deal with a scenario in which there is no pre-defined set of tags. 
\item We show state-of-the-art results on DUT-WEBV, a large dataset for tag localization in web videos. Moreover, we report an extensive experimental validation about the use of different web sources (Flickr, Google, Bing) to enrich and reinforce the video annotation.
\item We show how the proposed approach can be applied in a real-world scenario to perform open vocabulary tag annotation. To evaluate the results, we collected more than 5,000 frames from 40 YouTube videos and three individuals to manually verify the annotation.
\end{itemize}

\section{Related work}\label{sec:related}

Probably the most important effort in semantic video annotation is TRECVID \cite{smeaton-2006}, an evaluation campaign with the goal to promote progress in content-based retrieval from digital video archives.
Recently, online videos have also attracted the attention of researchers \cite{ytcat-2010,ballan-2011,toderici-2011,reddy-2013,Nga-2014}, since millions of videos are available on the web and they include rich metadata such as title, comments and user tags.




\subsection{Tags at the video-level}\label{sec:related_video}
A vast amount of previous work has addressed the problem of online video tagging using a simple classification approach with multiple categories and classes. Siersdorfer \etal \cite{siersdorfer-2009} proposed a method that combines visual analysis and content redundancy, strongly present in social sharing websites, to improve the quality of annotations associated to online videos. They first detect the duplication and overlap between two videos, and then propagate the video-level tags using automatic tagging rules. Similarly Zhao \etal \cite{zhao-2010} investigated techniques which allow annotation of web videos from a data-driven perspective. Their system implements a tag recommendation algorithm that uses the tagging behaviors in the pool of retrieved near-duplicate videos.

A strong effort has been made to design effective methods for harvesting images and videos from the web to learn models of actions or events and use this knowledge to automatically annotate new videos. This idea follows similar successful approaches for image classification \cite{fergus-2005,grauman-2008,optimol} but it has been applied only for the particular case of single-label classification.
To this end, a first attempt has been made by Ulges \etal \cite{ulges-2010} who proposed to train a concept detection system on web videos from portals such as YouTube. 
A similar idea is presented in \cite{sclaroff-2009} in which images collected from the web are used to learn representations of human actions and then this knowledge is used to automatically annotate actions in unconstrained videos. A main drawback of these works is that they require training classifiers for each label, and this procedure does not scale very well, especially on the web.
Very recently, Kordumova \etal \cite{kordumova-2014} have also studied the problem of training detectors from social media, considering both image and video sources, obtaining state-of-the-art results in TRECVID 2013 and concluding that tagged images are preferable over tagged videos.


\subsection{Tags at the keyframe(or shot)-level}\label{sec:related_frame}
Several methods have recently been proposed for unsupervised spatio-temporal segmentation of unconstrained videos \cite{essa-2010,hartmann-2012,tang-2013}.
Hartmann \etal \cite{hartmann-2012} presented an object segmentation system applied to a large set of weakly and noisily tagged videos. They formulate this problem as learning weakly supervised classifiers for a set of independent spatio-temporal video segments in which the object seeds are refined using graphcut. Although this method shows promising results, the proposed system requires a high computational effort to process videos at a large scale. Similarly, Tang \etal~\cite{tang-2013} have addressed keyframe segmentation in YouTube videos using a weakly supervised approach to segment semantic objects. The proposed method exploits negative video segments (i.e.~those that are not related to the concept to be annotated) and their distance to the uncertain positive instances, based on the intuition that positive examples are less likely to be segments of the searched concept if they are near many negatives. Both these methods are able to classify each shot within the video either as coming from a particular concept (i.e.~tag) or not, and they provide a rough tag-to-region assignment.

The specific task of tag localization, i.e. transferring tags from the whole video to the keyframe or shot level, has been addressed by a few different research groups.
Wang \etal \cite{mwang-2012} proposed a method for event driven web video summarization by tag localization and key-shot mining. They first localize the tags that are associated with each video into its shots by adopting a multiple instance learning algorithm \cite{gli-icmr11}, treating a video as a bag and each shot as an instance. Then a set of keyshots are identified by performing near-duplicate keyframe detection.
Zhu \etal \cite{zhu-2013} used a similar approach in which video tags are assigned to video shots analyzing the correlation between each shot and the videos in a corpus, using a variation of sparse group lasso.
A strong effort in collecting a standard benchmark for video localization research has been recently done by Li \etal \cite{dutwebv}. They released a public dataset designed for tag localization, composed by $1550$ videos collected from YouTube with 31 concepts and providing precise time annotations for each concept. The authors provide also an annotation baseline obtained using multiple instance learning, following \cite{gli-icmr11}.
All of these techniques have been largely adopted training classifiers, but still strongly suffer the lack of comprehensive, large-scale training data.

An early version of the proposed approach was introduced in our preliminary conference papers \cite{wsm-2010,ballan-2011}. In this paper we made key modifications in the algorithm and obtained significant improvements in the results. Differently from our previous work we introduce multiple types of image sources for a more effective cross-media tag transfer; we design a vote weighting procedure based on visual similarity and the use of a temporal smoothing strategy which exploits the temporal continuity of a video; further, we show a better performance in terms both of precision and recall. Finally, large-scale experiments have been carried on using a new public dataset \cite{dutwebv,hli-2014}, allowing fair comparisons w.r.t.~other methods.

\begin{figure*}[!t]
\centering
\includegraphics[width=1\textwidth]{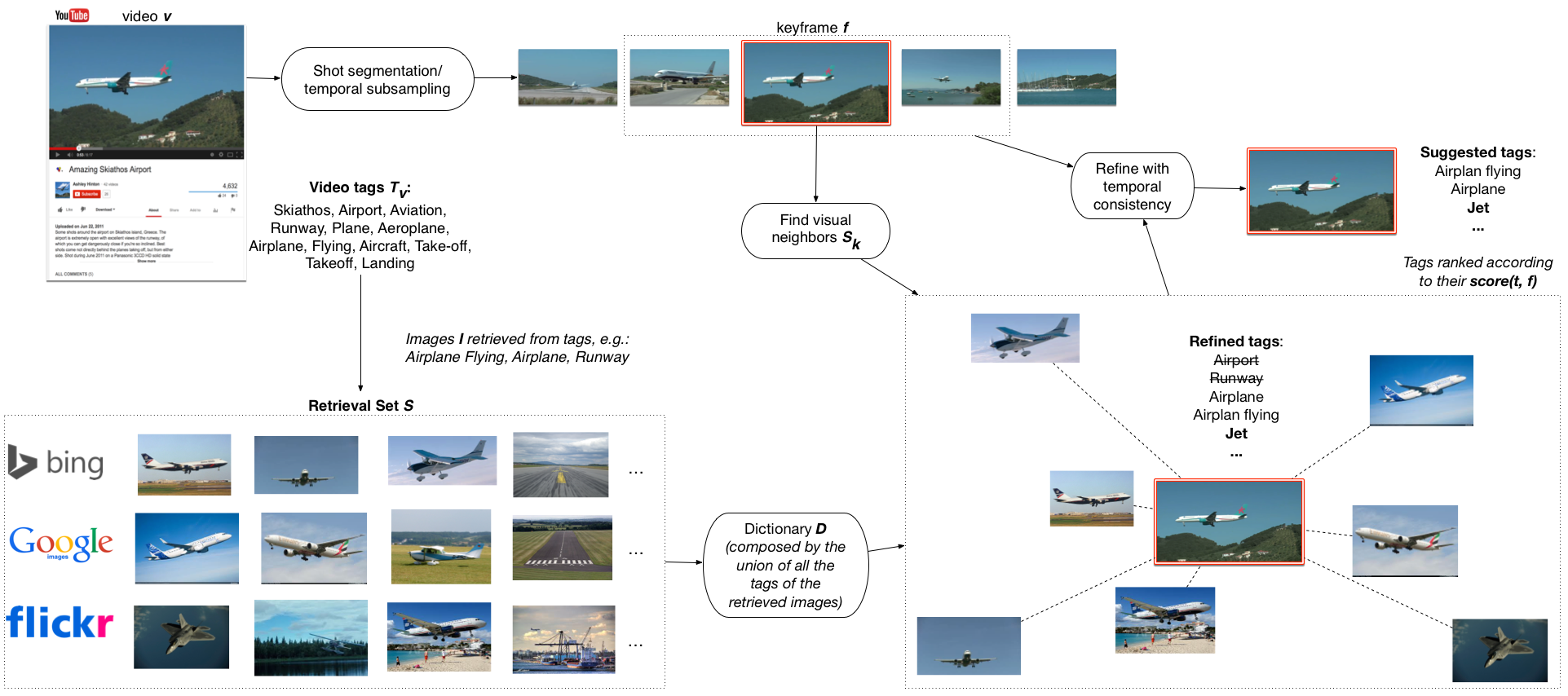}
\caption{Overview of the proposed method.}
\label{fig:system-schema}
\end{figure*}

\begin{table}[!h]
\centering
\textcolor{red}{
\begin{tabular}{p{1cm}p{7cm}} 
\toprule
\textbf{Variable} & \textbf{Meaning}	\\
\hline
$V$	& collection of videos and metadata (titles, tags, descriptions, etc.) \\
$v,f$ & a video from $D$ and a keyframe within $v$\\
$D$	& dictionary of tags to be used for annotation \\
$T_v,T'_v$ & set of tags associated to the video $v$, prior and after the Tag refinement and localization procedure \\
$S$ & set of images downloaded from Google, Bing and Flickr using $T_v$ filtered by stopwords, dates, tags containing numbers, punctuations and symbols \\
$t$ & a particular tag from $D$\\
$S_t$ & set of images from $S$ annotated with the tag $t$\\
$I_i,T_i$ & an image from $S$ and their tags\\
$S_K,T_K$ & set of $K$ image neighbors for a given keyframe $f$ ($S_K \subseteq S$) and their tags\\
$T_f$ & set of tags associated to the keyframe $f$ \\
$f^{(k)}$ & a keyframe at time $k$ \\
$t^{(k)}$ & a binary variable that defines whether the tag $t$ is present in the keyframe $f^{(k)}$ \\
\bottomrule
\end{tabular}}
\caption{Summary of notations used in this paper.}
\label{tab:notations}
\end{table}

\section{Approach}\label{sec:approach}

\textcolor{red}{The architecture of our system is schematically illustrated in Fig.~\ref{fig:system-schema} and our notation is defined in Tab.~\ref{tab:notations}.}
Let us consider a corpus $V$ composed of videos and metadata (e.g.~titles, tags, descriptions). We further define $D$ as a dictionary of tags to be used for annotation. Each video $v \in V$, with tags $T_v \subseteq D$, can be decomposed in different keyframes.

Online video annotation is performed in two stages: in the first stage a relevance measure of the video tags is computed for each keyframe, possibly eliminating tags that are not relevant; then new tags are added to the original list. Video keyframes can be obtained either from a video segmentation process or from a simple temporal frame subsampling scheme.
Each keyframe of the video is annotated using a data-driven approach, meaning that (almost) no training takes place offline. Given a keyframe, our method retrieves images from several sources and proceeds to transfer labels across similar samples.

\subsection{Retrieval set}
Similarly to several other data-driven methods \cite{makadia-2008,hays-2008,vicente-2011,liu-2011,spampinato-2015}, we first find a training set of tagged images that will serve for label propagation.
The tags $T_{v}=\{t_{1},\dotsc,t_{l}\}$ associated to a video $v$ are filtered to eliminate stopwords, dates, tags containing numbers, punctuations and symbols. 
In addition, we also include the WordNet synonyms of all these labels to extend the initial set of tags~\footnote{To cope with the fact that WordNet synonyms may introduce a semantic drift, for these tags we downloaded a number of images equal to one third of the original set.}. This resulting list of tags is then used to download a set of images $S=\{I_{1},\dotsc,I_{m} \}$ from Google, Bing and Flickr.
Following this procedure an image $I_{i} \in S$, retrieved using $t_j$ as query, has the following set of tags $T_i=\{t_{j}, t^{'}_{1},\dotsc,t^{'}_{z} \}$ if it has been obtained from Flickr or $T_i=\{t_{j}\}$ if it has been obtained from Google or Bing.
It has to be noticed that in the latter case, only the query term has been collected as a label since the images do not contain any other additional tag.
So let $D \supseteq T_{v}$ be the union of all the tags of the $m$ images in $S$, after that they have been filtered with the same approach used for video tags (i.e. removing the stopwords, dates etc.). This set $D$ is then used in the following steps to annotate ``on the fly'' the video.

Given the retrieval set $S$, for each keyframe $f$ within the video $v$ we find a (relatively) small set of $K$ visual neighbors $S_K \subseteq S$. A good neighbor set will contain images that have similar scene types or objects (in our experiments we varied $K$ from $150$ to $300$ images). In the attempt to indirectly capture this kind of similarity, we compute a 2000-d bag-of-visual-words descriptor, computed from densely sampled SIFT points. This descriptor can be efficiently used to find similar images using approximate search data structures by hierarchical k-means trees \cite{muja-2014}, in order to address scalability issues.

\subsection{Tag localization and refinement}
A simple approach to annotate a keyframe $f$ is to consider only the tags belonging to the set of tags $T_v$ that is associated to the video, computing their rank according to their relevance w.r.t.~the keyframe to be annotated. This is a common procedure used for image tagging \cite{makadia-2008,li-2009}.
However, this approach does not yield good results for the task of video annotation since the video tags may be associated only to certain keyframes and not to others. In fact, if we consider all the $t \in T_{v}$ for each keyframe, this procedure would simply result in a re-ranked list of the original video tags. 

In order to solve this problem, we adopt the following approach: a tag $t$ is kept in the list $T_{f}$, i.e.~the set of tags associated to the keyframe $f$, only if it is present among the tags of the visual neighborhood (noted as $T_K$).
Since the visual neighbors are images tagged by amateurs, such as Flickr users, or obtained from sources that can not be fully trusted, such as the images retrieved from Google or Bing, it is fundamental to evaluate the relevance of the tags that compose the lexicon.
To this end, we build on the tag relevance algorithm for social image retrieval by Li \etal \cite{li-2009}, and we present an effective framework to tackle the problem of tag localization and refinement in web videos.

The original \emph{tag relevance} algorithm is based on the consideration that if different persons label visually similar images using the same tags, then these tags are more likely to reflect objective aspects of the visual content.
Therefore it can be assumed that the more frequently the tag occurs in the neighborhood, the more relevant it might be. However, some frequently occurring tags are unlikely to be relevant for the majority of images.
To consider this fact, given a keyframe $f$, the tag relevance score takes into account a prior term obtained by computing the ratio of cardinality of images tagged with $t$ (denoted as $S_t$), to that of the entire retrieval set $S$:

\begin{equation}
tagRelevance(t,f,T_K):=\frac{1}{K}\sum_{i=1}^{K} R(t,T_i) - \frac{|S_t|}{|S|}
\label{eq:relevance}
\end{equation}

where 
\begin{equation}
R(t,T_i) = 
\begin{cases} 
1  & \mbox{if } t \in T_i \\ 
0  & otherwise 
\end{cases}
\label{eq:votes}
\end{equation}

\noindent where $| \cdot |$ is the cardinality of a set. Eq.~\ref{eq:relevance} shows that more neighbor images labeled with the tag $t$ imply larger tag relevance score. At the same time common frequent tags, that are less descriptive, are suppressed by the second term. 
\color[rgb]{0,0,0}

Differently from \cite{li-2009}, the $tagRelevance$ is not forced to be $\geq 1$ and in case no visual neighbor is associated to $t$ then it is set to $0$. This effectively allows to localize in time the original video tags.
	
The function $R(t,T_i)$ can be changed to account for the similarity between a keyframe and its visual neighbors.
In our system we weight each vote with the inverse of the square of Euclidean distance between $f$ and its neighbors:

\begin{equation}
R(t,T_i) = 
\begin{cases} 
\frac{1}{d( f, I_i)^2}  & \mbox{if } t \in T_i \\ 
0  & otherwise 
\end{cases}
\label{eq:wvotes}
\end{equation}

\noindent where $d( f, I_i)$ is the Eucliden distance between feature vectors of the keyframe $f$ and the image $I_i$. 
It has to be noticed that in case that a relevant tag is incorrectly eliminated in this phase, it may be recovered during the following stage of annotation.

Summarizing the above, the output of tag relevance estimation is a ranked list of tags for each keyframe $f$. In other words, $\forall t \in T_K$, the algorithm computes its tag relevance and a resulting rank position $rank_{t}$.
Then, for each tag in $T_f$ (as obtained from the previous steps), we compute the co-occurrence with all the tags in $T_K$. This results in a tag candidate list from which we select the tags that have a co-occurrence value that is above the average.
For each candidate tag we then compute a suggestion score $score(t,T_{f})$, according to the $Vote^{+}$ algorithm \cite{zwol-2008}. The final score is computed as follows: 
\begin{equation}
score(t,f) = score(t, T_{f}) \cdot \frac{\lambda}{\lambda+(rank_{t}-1)}
\label{eq:score}\end{equation}
where $\lambda$ is a damping parameter set to 20. We tuned $\lambda$ on our training set by performing a parameter sweep and maximizing performance both in terms of precision and recall. 
The results obtained applying Eq.~\ref{eq:score} are used to order all the candidate tags for the actual keyframe $f$, and the 5 most relevant tags are then selected.
Finally, the union of all the tags selected at the keyframe level may be used to annotate the video at the global level (hereafter referred as to $T'_v$).

\begin{algorithm}[!t]
\BlankLine
\KwIn{A test video $v$ with tags $T_v$.}
\KwOut{A set of keyframes $f \in v$ annotated with tags in~$D$, The refined set $T'_v$ at the video level.}
\BlankLine
Retrieve images from Google, Bing and Flickr for each $t \in T_v$ and let $S$ be the retrieval set while $D$ is the union of all the tags of the images in $S$\;
\BlankLine
\For{each keyframe $f \in v$}{
	Find $K$ nearest visual neighbors of $f$ from $S$\;
	\BlankLine
	$tagRelevance(t,f,T_K):=\sum_{i=1}^{K} R(t,T_i) - Prior(t,D)$\;
	\BlankLine
	Rank each candidate tag $t$ by $tagRelevance$ in descending order, and compute $score(t,f)$ (Eq.~\ref{eq:score})\;
	\BlankLine
	Refine/compute the final $score(t,f)$ by exploiting temporal continuity (Eq.~\ref{eq:temporal})\;
}
\BlankLine
Define $T'_v:=\bigcup_f T_f$ as the refined set of tags for $v$\;
\BlankLine
\caption{Tag refinement and localization}\label{algo}
\end{algorithm}

\subsection{Temporal consistency}\label{sec:temporal-consistency}

A main drawback of the procedure reported above is that the score computed using Eq.~\ref{eq:score} does not account for the temporal aspects of a video.
On the other hand, videos exhibit a strong temporal continuity in both visual content and semantics \cite{liu-2008}.
Thus we attempt to exploit this coherence by introducing a temporal smoothing to the relevance scores with respect to a tag.
To this end, for each tag $t$ and keyframe $f$, we re-evaluate the score function as reported below.

Let $f^{(k)}$ (or, for simplicity, $f$) be the actual keyframe at time $k$, and $d$ the maximum temporal distance within which the keyframes are considered; thus $f^{(k-i)}$ refers to the nearby keyframe at a temporal distance $i$. The score is computed as follows:

\begin{equation}
score(t,f)=\sum_{i=-d}^{+d} w_i \cdot P(t^{(k)}=1 | t^{(k-i)}=1) \cdot score(t,f^{(k-i)}).
\label{eq:temporal}
\end{equation}

The term $score(t,f^{(k-i)})$ is the score obtained for the tag $t$ and the keyframe that is temporally $i$ keyframes apart from $f$, while $w_i$ is a Gaussian weighting coefficient (which satisfies $\sum_{i}w_{i} = 1$).

The binary random variable $t^{(k)}$ is similarly defined to represent whether the tag $t$ is present in the keyframe $f^{(k)}$. We then estimate the conditional probabilities between neighboring keyframes (for a tag at a time), from ground-truth annotations.
These are computed as follows: 

\begin{equation}
P(t^{(k)}=1|t^{(k-i)}=1)=\frac{\#(t^{(k)}=1,t^{(k-i)}=1)}{\#(t^{(k-i)}=1)}
\end{equation}

\noindent where $\#(t^{(k-i)}=1)$ is equivalent to the total numbers of relevant keyframes in the training dataset; $\#(t^{(k)}=1,t^{(k-i)}=1)$ is the total number that two keyframes are $i$ frames apart and both relevant to the tag $t$. 

We examine the contributions of changing the width of time window $d$ in the experiments of Section \ref{sec:time_window}.
We finally summarize the procedure for tag refinement and localization by neighbor voting in Algorithm~\ref{algo}.

\section{Experiments}\label{sec:experiments}

Our proposed approach is a generic framework that can be used to annotate web videos and also to refine and localize their initial set of tags. To quantitatively evaluate the performance of our system, we present extensive experimental results for tag refinement and localization on a large public dataset.

\begin{figure}[!t]
\begin{tabular}{cccc}
\toprule
Video	&	Google	& Bing	& Flickr \\ \cmidrule{1-4}
\includegraphics[width=0.2\columnwidth]{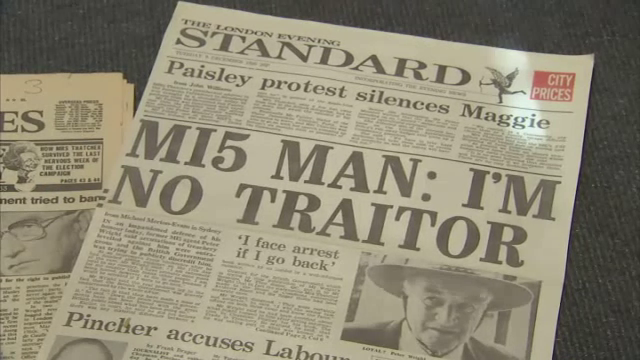}&\includegraphics[width=0.2\columnwidth]{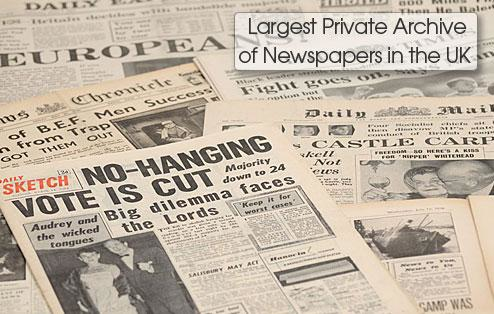}& \includegraphics[width=0.2\columnwidth]{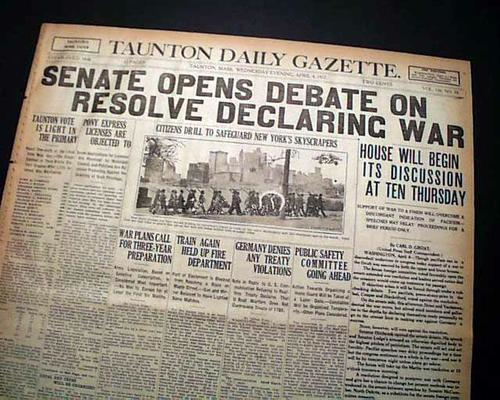}&\includegraphics[width=0.2\columnwidth]{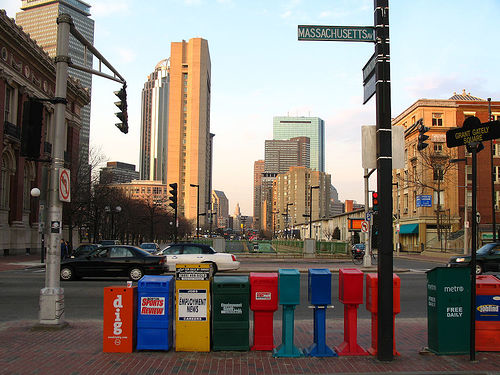}\\ \cmidrule{1-4}

\includegraphics[width=0.2\columnwidth]{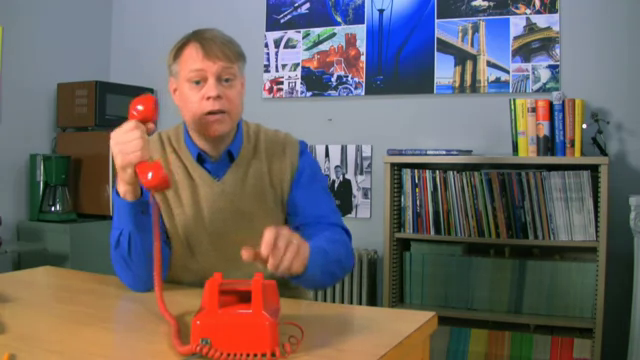}&\includegraphics[width=0.2\columnwidth]{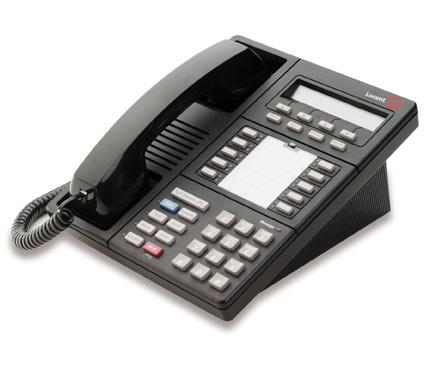}&\includegraphics[width=0.2\columnwidth]{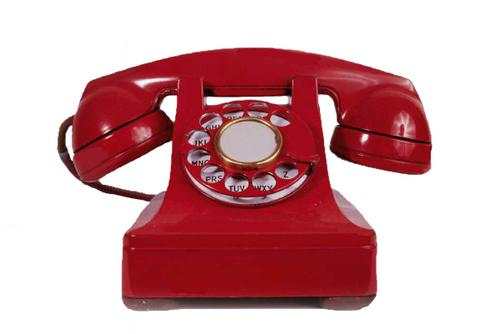}&\includegraphics[width=0.2\columnwidth]{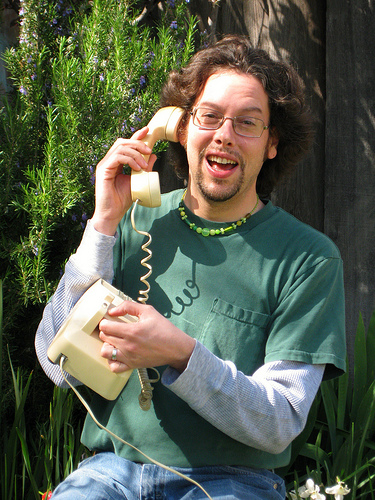}\\

\includegraphics[width=0.2\columnwidth]{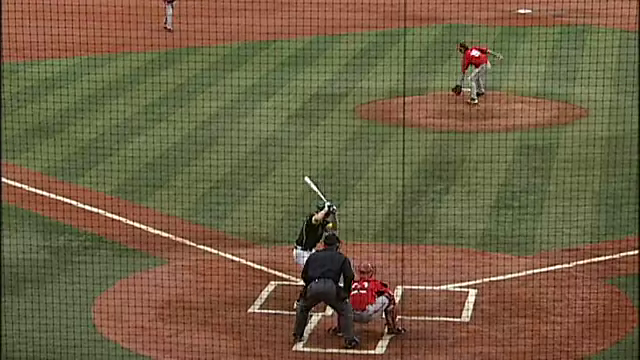}&\includegraphics[width=0.2\columnwidth]{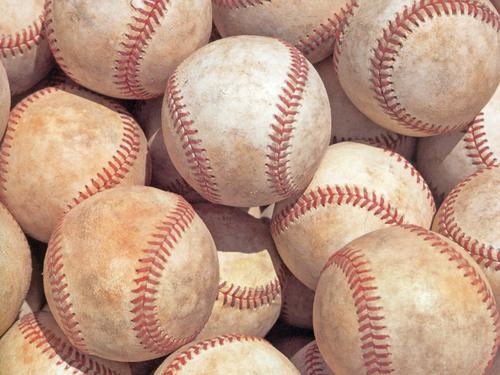}&\includegraphics[width=0.2\columnwidth]{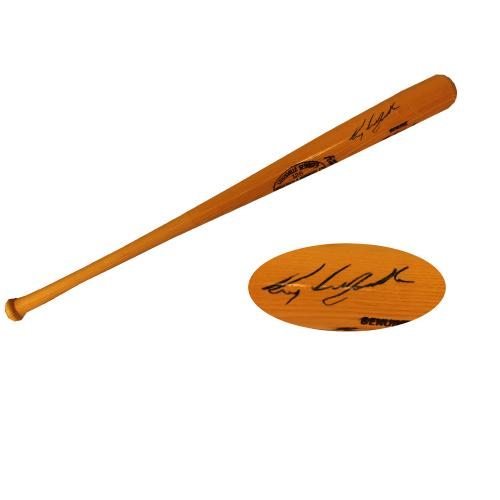}&\includegraphics[width=0.2\columnwidth]{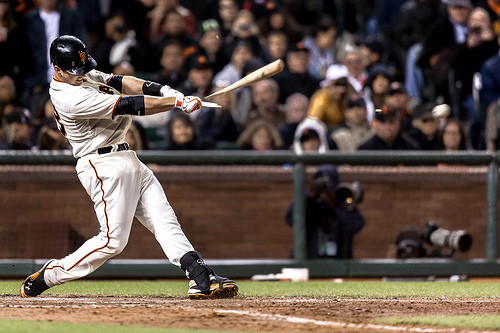}\\

\includegraphics[width=0.2\columnwidth]{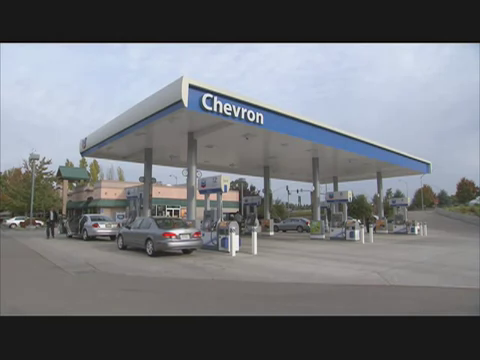}&\includegraphics[width=0.2\columnwidth]{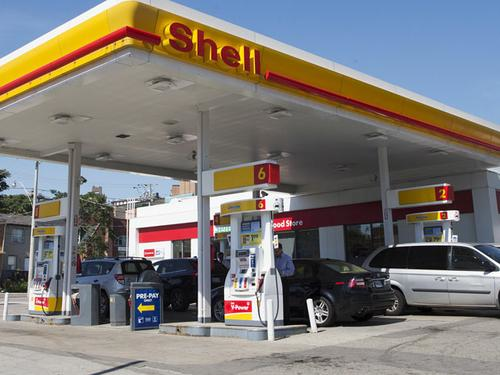}&\includegraphics[width=0.2\columnwidth]{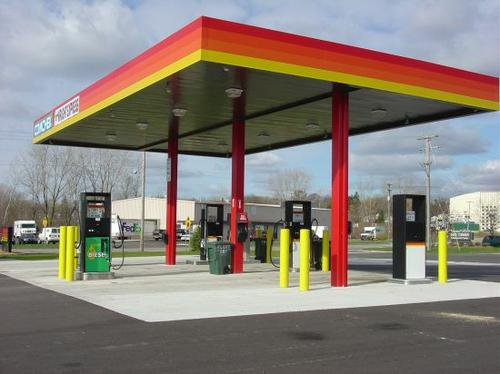}&\includegraphics[width=0.2\columnwidth]{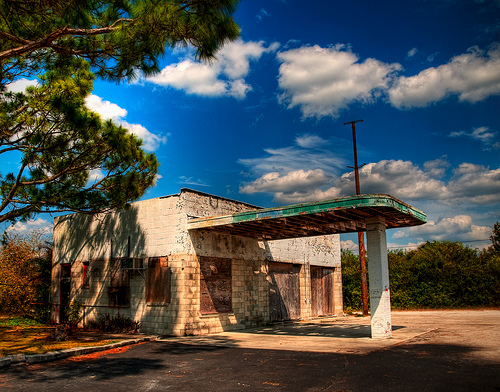}\\
\bottomrule
\end{tabular}
\caption{Example frames from YouTube videos, Google, Bing, Flickr images for the tag (top to bottom): \emph{newspapers}, \emph{telephones}, \emph{baseball} and \emph{gas station}.}
\label{fig:samples-sources}
\end{figure}

\subsection{DUT-WEBV dataset}\label{sec:dutwebv}
Our experiments have been conducted on the DUT-WEBV dataset \cite{dutwebv} which consists of a collection of web videos collected from YouTube by issuing 31 tags as queries. These tags, listed in Tab.~\ref{tab:concept-list}, have been selected from LSCOM \cite{lscom} and cover a wide range of semantic levels including \emph{scenes}, \emph{objects}, \emph{events}, \emph{people activities} and \emph{sites}.
There are $50$ videos for each concept, but $2$ videos are associated to two different tags, so that the total number of different videos is $1,458$. For each video is provided also a ground truth that indicates the time extent in which a particular tag is present.
In order to evaluate video annotation and tag refinement \dquote{in the wild}, we have collected additional information with respect to the original dataset. In particular, for each video that is still available on YouTube, we have extracted the tags provided by the original users to complement title and description that are provided by the authors of the dataset. This effort allows to use the dataset also for generic video annotation and tag refinement research, and it is so an additional contribution of our work.

\begin{table}[!t]
\begin{center}
\begin{tabular}{p{1.4cm}lp{1.5cm}p{1.7cm}}
\toprule
Category & Tag & \#frames \emph{with tag} & \#frames \quad \emph{total}\\ \bottomrule \cmidrule{1-4}
\multirow{5}{*}{Events}	& airplane flying	& 2,217	& 5,241 \\
					& birthday			&1,464	& 5,172 \\
					& explosion		& 2,050	& 3,870 \\
					& flood			& 2,216	& 4,083 \\
					& riot				& 4,462	& 6,582 \\
\cmidrule{1-4} 
\multirow{7}{*}{Objects}	& cows			& 3,014	& 5,080 \\
					& food			& 1,773	& 6,576 \\
					& golf player		& 1,497	& 4,295 \\
					& newspapers		& 2,443	& 6,168 \\
					& suits			& 2,287	& 5,302 \\
					& telephones		& 2,720	& 5,587 \\
					& truck			& 2,382	& 6,171 \\
\cmidrule{1-4} 		
\multirow{12}{*}{Activities}&	baseball	& 2,459	& 3,991 \\
					& basketball		& 3,026	& 4,925 \\
					& cheering		& 2,788	& 6,605 \\
					& dancing			& 1,781	& 6,092 \\
					& handshaking		& 1,516	& 3,412 \\
					& interviews		& 4,217	& 7,206 \\
					& parade			& 3,445	& 5,756 \\
					& running			& 2,826	& 6,024 \\
					& singing			& 4,045	& 6,802 \\
					& soccer			& 3,204	& 4,747 \\
					& swimming		& 2,757	& 4,924 \\
					& walking			& 2,669	& 6,035 \\
\cmidrule{1-4} 
\multirow{3}{*}{Scenes}	& beach			& 3,016	& 5,305 \\
					& forest			& 4,157	& 7,001 \\
					& mountain		& 2,735	& 6,394 \\
\cmidrule{1-4} 			
\multirow{4}{*}{Sites}		& aircraft cabin		& 2,593	& 5,110 \\
					& airport			& 4,187	& 6,538 \\
					& gas station		& 1,029	& 4,327 \\
					& highway			& 2,321	& 5,166 \\
\cmidrule{1-4} 			
\multirow{1}{*}{\emph{Total}}		& 		& \emph{83,296}	& \emph{170,302} \\					
\bottomrule
\end{tabular}
\caption{DUT-WEBV dataset: list of tags with their corresponding category, number of frames containing a particular tag/concept and total number of keyframes extracted from all the videos labeled with a particular tag.\vspace{-10pt}}\label{tab:concept-list}
\end{center}
\end{table}%

Our experimental setup follows the one proposed by the authors of the dataset, whose results are compared in Sect.~\ref{sec:baseline-comparison}. Video frames have been sub-sampled from each video every two seconds, following the experimental setup proposed by the authors of the dataset, obtaining $170,302$ different frames.
For each tag we have obtained images from web search engines, namely Google Images and Bing Images, and from a social network, i.e.~Flickr. The overall number of images retrieved is $61,331$.
Considering all the video frames and downloaded images, the overall number of images in the dataset is thus $231,633$, comparable to the dimension of NUS-WIDE (which is nowadays the largest common dataset used for social image retrieval and annotation).
Some examples of the images retrieved from these web sources, as well as the corresponding keyframes from DUT-WEBV, are shown in Fig.~\ref{fig:samples-sources}.

\begin{table*}
\begin{center}
\begin{tabular}{ccccccccccccc}
\toprule
\multirow{2}{*}{\#num.~neigh.}		& \multicolumn{6}{c}{$Precision@1$}	& \multicolumn{6}{c}{$Recall@1$} 	\\ 
\cmidrule(lr){2-7}\cmidrule(lr){8-13} & events	& objects	& activities	& scenes	& sites	& \textbf{Avg.} 	& events	& objects	& activities	& scenes	& sites	& \textbf{Avg.}  \\ \cmidrule{1-13}\morecmidrules\cmidrule{1-13}
150 			& 59.4	& 52.9	& 60.7		& 70.8	& 50.5	& \textbf{58.4} 	& 54.4	& 46.3	& 41.4		& 55.5	& 66.6	& \textbf{49.2} \\ \cmidrule{1-1}\cmidrule(lr){2-7}\cmidrule(lr){8-13}
200 			& 59.8	& 54.2	& 59.8		& 70.9	& 52.4	& \textbf{58.6} 	& 53.3	& 48.7	& 40.6		& 55.9	& 68.8	& \textbf{49.6} \\ \cmidrule{1-1}\cmidrule(lr){2-7}\cmidrule(lr){8-13}
250 			& 57.7	& 53.4	& 60.9		& 70.4	& 53.1	& \textbf{58.6} 	& 53.2	& 48.4	& 42.4		& 54.4	& 69.1	& \textbf{50.1} \\ \cmidrule{1-1}\cmidrule(lr){2-7}\cmidrule(lr){8-13}
300 			& 59.0	& 54.7	& 62.3		& 69.5	& 53.3	& \textbf{59.6} 	& 54.8	& 48.6	& 42.9		& 55.1	& 71.4	& \textbf{50.9} \\ 
\bottomrule
\end{tabular}
\caption{Results of tag localization using only DUT-WEBV data (Experiment 1).}\label{tab:exp1}
\end{center}
\end{table*}%

\subsection{Experiment 1: tag localization using only DUT-WEBV data} 
First of all, we present a tag localization baseline on the DUT-WEBV dataset relying only on the keyframes extracted from the web videos. The experimental setup used to build the image retrieval set follows the approach used in the baseline provided with the dataset \cite{dutwebv}. So, given a particular tag $t$ to be localized in a video $v$, we extract all the keyframes of the other videos associated to $t$, and the keyframes of $10$ randomly selected videos associated to other $10$ randomly selected tags from $T_v$. Similarly to previous works \cite{dutwebv,hli-2014}, we use $Precision@N$ and $Recall@N$ to evaluate results (i.e. precision/recall at top $N$ ranked results).

In our experiments, the visual neighborhood $S_K$ is obtained varying the number $K$ of neighbors from $150$ to $300$. Tag relevance is computed using Eq.~\ref{eq:score} and without weighting votes.
These preliminary results are reported in Tab.~\ref{tab:exp1}. It can be observed that, as the number of visual neighbors increases so the performance slightly improves, both in terms of precision and recall. In the rest of the paper, if not mentioned otherwise, we fixed $K=200$.
We have conducted also similar experiments by weighting votes as reported in Eq.~\ref{eq:wvotes}. Using this procedure we observed an improvement in recall of around $4\%$ and a loss in precision of more then $5\%$. The tag localization task is inherently more demanding in terms of precision since a tag that has not been recognized at a particular keyframe might be recovered in the forthcoming frames. So, in the following experiments, we only report the performance obtained using the original voting scheme (Eq.~\ref{eq:votes}).

\subsection{Experiment 2: tag localization using different web sources}\label{sec:exp-sources}
In this experiment we evaluate the effect of using different sources to build the visual neighborhood.
First of all we compare the results obtained with the previous baseline configuration (i.e. video only) with several combination of video and different web sources. Then we analyze the same configurations without the original video frames. 
Note that in these experiments the diversity of the images in the retrieval set grows, as well as the total number of tags in our dataset.
The results are reported in Tab.~\ref{tab:exp4}; the first column indicates the sources used to create the neighborhood.

\begin{figure*}[!ht]
\centering
\includegraphics[width=1\textwidth]{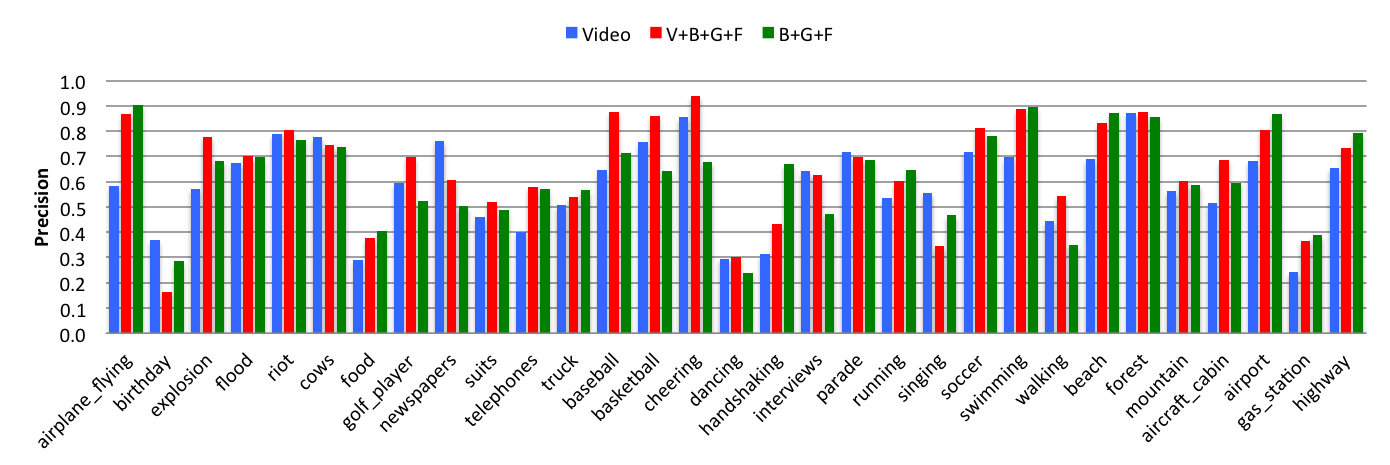}
\caption{Precision rate broken down by tag for the most interesting combinations of image sources.}
\label{fig:alltags}
\end{figure*}

It can be observed that using all the available image sources provides the best precision result of $65.2\%$.
In terms of precision any combination of video and additional source performs better than the same source alone, but it is interesting to notice that using all the social and web sources together (B+G+F) provides very good results, $62.3\%$.
This is even better than using video alone, which achieves $58.6\%$, 
or any combination of video with a single additional image source~\footnote{We believe the main difference between the use of Google and Bing is due to technical reasons: these search engines do not provide an official API to download the images needed for the experiments, that were obtained from them through scraping. Bing apparently enforces stricter anti-scraping techniques that resulted in a more limited and less diverse set of images than Google.} except when using Flickr. 
We believe that the results obtained using only web sources (B+G+F) are very interesting since this configuration might be the most useful in a \dquote{annotation in the wild} scenario, in which no previous video data are available. It has also to be noticed that this configuration provides higher results w.r.t.~the \dquote{closed world} scenario in which only video data is used, on almost all the categories.
In some cases, look for example at tags such as \emph{highway}, \emph{airport} and \emph{airplane\_flying}, the performance are significantly higher than in the baseline configuration.
A comparison of the precision obtained for each individual tag with the most interesting configurations is shown in Fig.~\ref{fig:alltags}.

Regarding recall results, the main difference is between using only video data which achieves $49.6\%$ and any other combination which provides at most $29.9\%$.
In case of using video alone, we rely on the training data provided in the original benchmark and this is obviously not possible in a real application of our system in which the set of tags is not known a priori.
Moreover, the intra-class variation of the videos is not very high and this may facilitate too much the recall results.

\begin{table*}[!htbp]
\begin{center}
\begin{tabular}{lcccccccccccc}
\toprule
\multirow{2}{*}{sources}			& \multicolumn{6}{c}{$Precision@1$}	& \multicolumn{6}{c}{$Recall@1$} 	\\ 
\cmidrule(lr){2-7}\cmidrule(lr){8-13}	& events	& objects	& activities	& scenes	& sites	& \textbf{Avg.} 	& events	& objects	& activities	& scenes	& sites	& \textbf{Avg.}  \\ \cmidrule{1-13}\morecmidrules\cmidrule{1-13}
V & 59.8	& 54.2	& 59.8	& 70.9	& 52.4	& \textbf{58.6}	& 53.3	& 48.7	& 40.6	& 55.9	& 68.8	& \textbf{49.6}   \\ \cmidrule{1-1}\cmidrule(lr){2-7}\cmidrule(lr){8-13}
V + B + G + F	& 66.3	& 58.1	& 66.0	& 77.0	& 64.7	& \textbf{65.2}	& 27.3	& 28.4	& 20.7	& 43.7	& 24.3	& \textbf{26.2}   \\ \cmidrule{1-1}\cmidrule(lr){2-7}\cmidrule(lr){8-13}
V + B + G	& 69.9	& 54.1	& 68.2	& 76.7	& 62.3	& \textbf{65.1}	& 25.3	& 26.1	& 18.3	& 41.8	& 26.4	& \textbf{24.5}   \\ \cmidrule{1-1}\cmidrule(lr){2-7}\cmidrule(lr){8-13}
V + B	& 65.3	& 56.7	& 60.9	& 72.5	& 56.9	& \textbf{61.3}	& 25.9	& 32.3	& 23.1	& 51.1	& 35.0	& \textbf{29.9}   \\ \cmidrule{1-1}\cmidrule(lr){2-7}\cmidrule(lr){8-13}
V + G	& 81.2	& 53.1	& 54.5	& 62.2	& 72.6	& \textbf{62.1}	& 48.1	& 44.0	& 5.2	& 27.5	& 43.1	& \textbf{29.3}   \\ \cmidrule{1-1}\cmidrule(lr){2-7}\cmidrule(lr){8-13}
V + F	& 70.4	& 54.7	& 64.7	& 77.1	& 54.7	& \textbf{63.3}	& 25.2	& 29.2	& 20.4	& 34.1	& 18.8	& \textbf{24.3}   \\ \cmidrule{1-1}\cmidrule(lr){2-7}\cmidrule(lr){8-13}
B + G + F	& 66.6	& 54.2	& 60.3	& 77.1	& 66.0	& \textbf{62.3}	& 24.9	& 29.4	& 13.5	& 35.1	& 18.8	& \textbf{21.7}   \\ \cmidrule{1-1}\cmidrule(lr){2-7}\cmidrule(lr){8-13}
B + G	& 63.7	& 54.1	& 51.4	& 76.2	& 64.4	& \textbf{58.1}	& 25.6	& 24.4	& 11.5	& 37.6	& 20.8	& \textbf{20.4}   \\ \cmidrule{1-1}\cmidrule(lr){2-7}\cmidrule(lr){8-13}
B	& 47.1	& 54.7	& 57.2	& 66.1	& 21.6	& \textbf{51.3}	& 4.4	& 23.6	& 12.3	& 38.5	& 1.4	& \textbf{14.7}   \\ \cmidrule{1-1}\cmidrule(lr){2-7}\cmidrule(lr){8-13}
G	& 64.8	& 52.9	& 54.8	& 75.7	& 61.4	& \textbf{58.9}	& 31.8	& 22.4	& 11.0	& 28.8	& 23.0	& \textbf{20.2}   \\ \cmidrule{1-1}\cmidrule(lr){2-7}\cmidrule(lr){8-13}
F	& 51.2	& 43.1	& 56.4	& 71.6	& 44.5	& \textbf{52.5}	& 19.0	& 27.1	& 15.1	& 8.9	& 4.9	& \textbf{16.5}   \\ 
\bottomrule
\end{tabular}
\caption{Results of tag localization using different combinations of image sources: DUT-WEBV frames (V), social images from Flickr (F) and web images from Google (G) and Bing (B). The visual neighborhood consists of 200 images.}\label{tab:exp4}
\end{center}
\end{table*}

\begin{table}[!ht]
\begin{center}
\begin{tabular}{p{1.6cm}rrrr}
\toprule
Tag 		& Video & Google& Bing	& Flickr\\ \cmidrule{1-5}\morecmidrules\cmidrule{1-5}
newspapers	& 76.2	& 81.3	& 90.4	& 45.2 \\ \cmidrule{1-5} 
telephones	& 40.2	& 36.7	& 40.5	& 47.1 \\ \cmidrule{1-5} 
baseball	& 64.5	& 48.5  & 62.2  & 83.1	\\ \cmidrule{1-5}
gas station & 24.2	& 37.5	& 33.8	& 0.0	\\  		
\bottomrule
\end{tabular}
\caption{$Precision@1$ for specific tags, when using different image sources.}\label{tab:analysis-tags}
\end{center}
\end{table}

\paragraph{Analysis of specific tags}
To analyze more in depth the effect of using different image sources, the following Tab.~\ref{tab:analysis-tags} reports the results for a few tags that have large variations in terms of precision when using only one of the possible sources.
Some of these variations can be motivated by the fact that images of some social sources, such as Flickr, have been created with a different intent. For example, on the one hand Flickr is often used by amateur photographers aiming at some aesthetics, and are therefore too different from videos that are documenting an object, like newspapers of gas stations.
On the other hand Flickr users tend to represent objects like telephones or activities like baseball in their context, i.e.~including persons using them or participating in the action, while Bing and Google tend to use more objective images. Examples of these differences are shown in Fig.~\ref{fig:samples-sources}.

\subsection{Experiment 3: tag localization using temporal consistency}\label{sec:time_window}
In this experiment we evaluate the effect of our temporal smoothing procedure (see Sect.~\ref{sec:temporal-consistency}) using the combination of parameters obtained from previous experiments that obtains the best precision, i.e.~using all image sources and $K=200$.
In Tab.~\ref{tab:exp5} we show the results obtained at varying width of keyframe time window, i.e.~the value of $d$ in Eq.~\ref{eq:temporal}. Keyframes have been temporally subsampled every 2 seconds, therefore if $d=1$ the temporal extent of the video corresponds to 4 seconds.

The results show that considering temporal aspects is beneficial for the performance since it improves recall (around $4\%$) without reducing precision.
Using larger temporal extents does not provide particular advantages since conditional probabilities of the presence of a concept at several seconds of distance are often not relevant. It has to be noticed that our temporal smoothing procedure has a negligible computational cost and so it gives great advantages with no drawbacks.

\begin{table*}[!htbp]
\begin{center}
\begin{tabular}{ccccccccccccc}
\toprule
\multicolumn{7}{c}{Precision@1}	& \multicolumn{6}{c}{Recall@1} 	\\ 
\cmidrule{1-1}\cmidrule(lr){2-7}\cmidrule(lr){8-13}	d	& events	& objects	& activities	& scenes	& sites	& \textbf{Avg.} 	& events	& objects	& activities	& scenes	& sites	& \textbf{Avg.}  \\ \cmidrule{1-13}\morecmidrules\cmidrule{1-13}
0 		& 66.3	& 58.1	& 66.0		& 77.0	& 64.7	& \textbf{65.2}	& 27.3	& 28.4	& 20.7		& 43.7	& 24.3	& \textbf{26.2} \\ \cmidrule{1-1}\cmidrule(lr){2-7}\cmidrule(lr){8-13}
1		& 64.4	& 57.9	& 66.5		& 77.0	& 66.9	& \textbf{65.3}	& 31.2	& 29.8	& 22.5		& 48.3	& 26.7	& \textbf{28.6} \\ \cmidrule{1-1}\cmidrule(lr){2-7}\cmidrule(lr){8-13}
2		& 64.9	& 57.7	& 66.0		& 76.6	& 67.0	& \textbf{65.1}	& 32.4	& 29.9	& 22.9		& 49.9	& 27.4	& \textbf{29.2} \\ \cmidrule{1-1}\cmidrule(lr){2-7}\cmidrule(lr){8-13}
3		& 64.8	& 57.8	& 66.1		& 76.6	& 67.5	& \textbf{65.3}	& 32.6	& 29.9	& 23.0		& 49.9	& 27.7	& \textbf{29.7} \\ 
\bottomrule
\end{tabular}
\caption{Results of tag localization with temporal consistency, using different time widths. Visual neighborhood obtained using $K=200$ from DUT-WEBV frames, social and web images (Flickr, Google Images, Bing Images). If $d$ is 0 then no temporal consistency is used.}\label{tab:exp5}
\end{center}
\end{table*}%

\subsection{Comparison with previous works}\label{sec:baseline-comparison}
The tag localization method proposed by the authors of the dataset is MIL-BPNET \cite{zhang-2004}. The choice of this method is motivated by the fact that MIL has been used in other approaches for tag localization \cite{gli-icmr11,mwang-2012,dutwebv} and thus provides a sound baseline.
In particular, given a tag $t$, the associated videos form the positive bags, while the others the negative bags. To reduce computational costs, for each tag, 10 negative tags are selected and for each negative tag 10 videos are randomly selected to create the negative bags.
Performance is reported by the original authors as $Precision@N$, with a varying $N$ that accounts for the number of video frames that contain each concept and the percentage of video parts that contains the concept.

We show in Tab.~\ref{tab:comparison} a comparison of the results reported in \cite{dutwebv} with our best combination of image sources (V+B+G+F) and temporal smoothing computed using $d=3$.
The table reports also figures of precision for two other methods as reported in \cite{hli-2014}: the first one use kernel density estimation (KDE) to localize the most relevant frames for a given tag; the second one combines KDE with visual topic modeling (using LDA). For these latter methods, the original authors report results for a subset of only ten tags.

Our proposed method obtains better results than MIL-BPNET for all tags but four, and overall performs better in all categories. On average, we outperform the baseline of $10\%$.
Moreover, a comparison w.r.t.~KDE and KDE+LDA shows that the proposed method obtains better results except for two tags.
It has to be noticed that our results are reported as $Precision@1$ while these baselines were measured using $Precision@N$ (with a large $N$), and so our improvements should be considered even more.

We compare also with a ConvNet-based classifier \cite{krizhevsky-2012}, trained using ImageNet 2010 metadata.
Very recently, deep convolutional neural networks (CNN) have demonstrated state-of-the-art results for large-scale image classification and object detection \cite{krizhevsky-2012,ilsvrc-2015} and promising results on multilabel image annotation \cite{gong-2014}.
Similarly to our method, results are reported as $Precision@1$. As can be expected convolutional neural networks have a better performance in several classes, although in many others the results are comparable (e.g.~basketball, soccer, gas station). On the other hand it has to be noted that even using a GPU implementation~\footnote{CCV -- A Modern Computer Vision Library: \url{http://libccv.org}} the processing time is twice as slower than the proposed method, and that using ConvNets require an extremely large amount of manually annotated images.

\begin{table}
\centering
\resizebox{1\columnwidth}{!}{
\begin{tabular}{l l c c c c c}
\toprule
Category& Tag			& \textbf{Our} 		& MIL  				&	KDE				& KDE +					& ConvNet\\
		&				& 					& \cite{zhang-2004}	& 	\cite{hli-2014}	& LDA \cite{hli-2014}	& \cite{krizhevsky-2012} \\
\cmidrule(lr){1-1} \cmidrule(lr){2-2} \cmidrule(lr){3-3} \cmidrule(lr){4-4} \cmidrule(lr){5-5} \cmidrule(lr){6-6} \cmidrule(lr){7-7}
\multirow{5}{*}{Events}	& airplane flying	&	84.3 & 72.6 	&	72.0			& 72.9 		& - \\
	&	birthday		&	12.7			&	30.5			&	-				& -		 	& - \\
	&	explosion		&	82.0			&	65.0			&	-				& -		 	& - \\
	&	flood			&	69.6			&	55.0			&	58.3			& 63.1	 	& - \\
	&	riot			&	78.8			&	69.3			&	-				& -		 	& - \\
	&	\textbf{Avg.}		&	\textbf{65.5}		&	\textbf{58.5} & \textbf{-} & \textbf{-} & \textbf{-} \\ \cmidrule(lr){1-7}
\multirow{7}{*}{Objects}	& cows	&	72.0	&	58.1			&	-				& -		 	& - \\
	&	food			&	37.3			&	41.6			&	-				& -			& - \\
	&	golf player		&	72.6			&	38.6			&	-				& -			& 61.8 \\
	&	newspapers		&	58.2			&	41.6			&	-				& -			& 64.3 \\
	&	suits			&	51.5			&	42.5			&	54.4			& 54.6		& 75.3 \\
	&	telephones		&	59.7			&	53.4			&	58.1			& 58.4		& 72.2 \\
	&	truck			&	53.3			&	52.1			&	-				& -			& 77.1 \\
	&	\textbf{Avg.}		&	\textbf{57.8}		&	\textbf{46.8} & \textbf{-} & \textbf{-} & \textbf{-} \\ \cmidrule(lr){1-7}
\multirow{12}{*}{Activities}&	baseball	&	91.2	&	66.9	&	-				& -			& 95.4 \\
	&	basketball		&	84.1			&	64.3			&	-				& -			& 87.9 \\
	&	cheering		&	96.3			&	58.2			&	-				& -		 	& - \\
	&	dancing			&	30.7			&	28.1			&	-				& -		 	& - \\
	&	handshaking		&	45.9			&	44.7			&	-				& -		 	& - \\
	&	interviews		&	71.2			&	61.8			&	65.6			& 69.6	 	& - \\
	&	parade			&	67.8			&	69.4			&	-				& -		 	& - \\
	&	running			&	62.3			&	45.5			&	47.0			& 54.7		& 34.2 \\
	&	singing			&	19.0			&	61.1			&	-				& -		 	& - \\
	&	soccer			&	82.6			&	76.3			&	71.4			& 79.7		& 83.8 \\
	&	swimming		&	86.5			&	70.8			&	-				& -			& 47.3 \\
	&	walking			&	55.2			&	43.0			&	-				& -		 	& - \\
	&	\textbf{Avg.}	&	\textbf{66.1}	&	\textbf{57.5} 	& \textbf{-} 		& \textbf{-} & \textbf{-} \\ \cmidrule(lr){1-7}
\multirow{3}{*}{Scenes}	& beach	&	85.0	&	70.5			&	-				& -		 	& - \\
	&	forest			&	83.5			&	73.2			&	76.3			& 79.5	 	& - \\
	&	mountain		&	61.6			&	57.4			&	53.9			& 58.6  	& - \\
	&	\textbf{Avg.}		&	\textbf{76.7}		&	\textbf{67.0} & \textbf{-} & \textbf{-} & \textbf{-} \\	\cmidrule(lr){1-7}
\multirow{4}{*}{Sites}	& aircraft cabin&	75.4	&	51.9	&	-				& -		 	& - \\
	&	airport			&	80.9			&	70.1			&	-				& -		 	& - \\
	&	gas station		&	41.1			&	23.5			&	-				& -			& 45.5 \\
	&	highway			&	72.9			&	58.5			& 	57.6			& 59.6  	& - \\
	&	\textbf{Avg.}		&	\textbf{67.6}		&	\textbf{51.0} & \textbf{-} & \textbf{-} & \textbf{-} \\		
\cmidrule(lr){1-6}\morecmidrules\cmidrule(lr){1-7}
	&	\textbf{Overall Avg.}&	\textbf{65.3}	&	\textbf{55.3} & \textbf{-} & \textbf{-} & \textbf{-} \\
\bottomrule
\end{tabular}}
\caption{Comparison between our method and the MIL-BPNET \cite{zhang-2004} baseline in terms of precision. We report also the results of KDE and KDE+LDA \cite{hli-2014} for a subset of nine tags as in the original paper.}\label{tab:comparison}
\end{table}

\subsection{Experiment 4: frame-level annotation ``in the wild''}
In this experiment we evaluate the performance of the annotation using an open vocabulary, performing tag localization and refinement. To this end we have selected 40 YouTube videos, for a total of $5,351$ frames; visual neighborhoods have been built from Google, Bing and Flickr images, retrieved using the original video tags. The dictionary used for annotation is obtained by the union of all the tags of the retrieved images, and on average is composed by around $8,000$ tags per video. With this approach it becomes possible to tag keyframes showing specific persons (e.g.~TV hosts like Ellen DeGeneres), objects (e.g.~Dodge Viper car or NBA Baller Beats videogame) or classes (e.g.~marathon races).
The annotation performance has been evaluated in terms of Precision@5 and Precision@10, through manual inspection of each annotated frame by three different persons, and averaging the results. Each annotator was requested to evaluate the relevance of each tag with respect to the visual content of each frame. Given the difficulty of this assessment this was performed after watching the whole video, and reading video title, description and list of original tags, so to understand the topics of each video and the content of the individual frames; frames were presented to the annotators following their order of appearance in the video.
Results are reported in Tab.~\ref{tab:wild-annotation}, comparing the results with a baseline that randomly selects tags, with a probability proportional to their frequency in the downloaded images. As can be expected the precision is lower than that of the other experiments, but this is due to the difficulty of multi-label annotation and to the very large vocabulary used to annotate each video.

\begin{table}[!h]
\centering
\begin{tabular}{lcc}
\toprule
Method		& Precision@5	& Precision@10	\\
\cmidrule(lr){1-1} \cmidrule(lr){2-2} \cmidrule(lr){3-3} 
Random       & 6.1  & 4.5  \\
\textbf{Our} & 33.4 & 30.4 \\
\bottomrule
\end{tabular}
\caption{Annotation ``in the wild'', using an open vocabulary. Comparison between our method and the random baseline.}
\label{tab:wild-annotation}
\end{table}

\begin{figure*}
\centering
\begin{tabular}{ccc}
\includegraphics[width=0.31\textwidth]{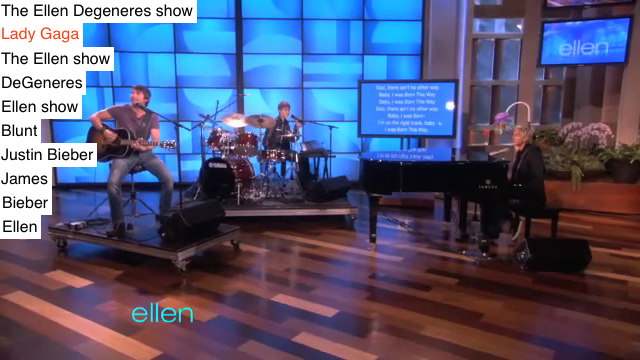}&\includegraphics[width=0.31\textwidth]{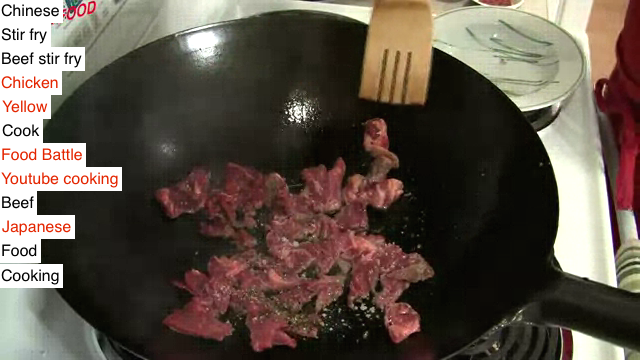}&\includegraphics[width=0.31\textwidth]{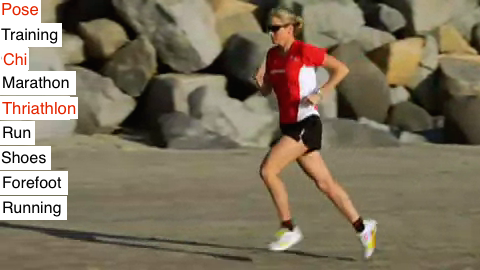}\\
\includegraphics[width=0.31\textwidth]{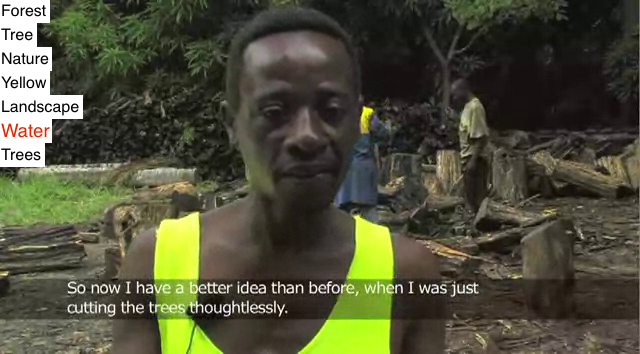}&\includegraphics[width=0.31\textwidth]{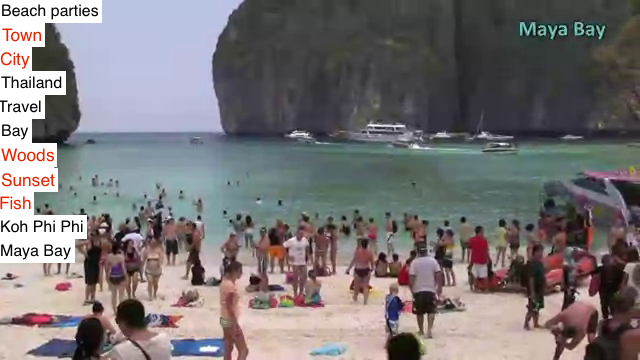}&\includegraphics[width=0.31\textwidth]{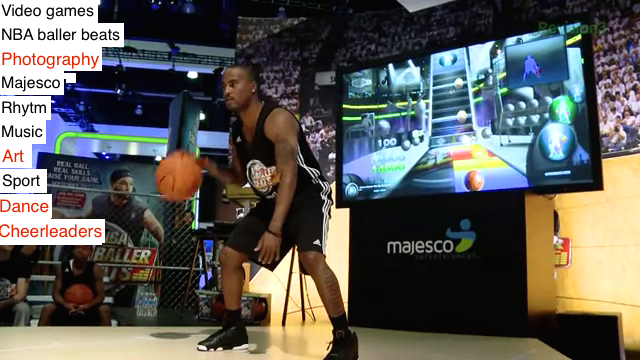}
\end{tabular}
\caption{Annotation ``in the wild'' experiment: some example frames from a subset of YouTube videos of the original DUT-WEBV dataset.}
\label{fig:annotation}
\end{figure*}

\subsection{Running time and system details}

Finally, we provide a rough analysis about the computational requirements of our system. \textcolor{red}{The Python implementation of the proposed algorithm annotates a video frame in about $0.17$ seconds, of which $96\%$ of the time is spent in computing the visual neighborhood, and $\sim4\%$ to compute tag localization and suggestion. 
The time required to compute temporal consistency is negligible. The average DUT-WEBV video is composed by around $110$ keyframes (with a median of 98), requiring about 18.7 seconds to process it.}
This is mostly an un-optimized and un-parallelized implementation, and all our experiments are run on a single workstation with Xeon $2.67$ GHz six core CPU and 48 GB RAM.
As previously reported, for each image and keyframe we have computed a 2000-d bag-of-visual-words histogram obtained from densely sampled SIFT descriptors. Moreover, we used ANN and hierarchical k-Means trees \cite{muja-2014} to speed up nearest neighbor search.

In order to promote further research on this topic, we provide all the additional annotation of the DUT-WEBV dataset to the public at large on our webpage \url{www.micc.unifi.it/vim}, as well as the visual features used in our experiments. We share also the images retrieved from the different web sources to build our retrieval set.

\section{Conclusions}\label{sec:conclusions}

In this paper we have presented a tag refinement and localization approach based on lazy learning. Our system exploits collective knowledge embedded in user generated tags and visual similarity of keyframes and images uploaded to social sites like YouTube and Flickr, as well as web image sources like Google and Bing.
We also improve our baseline algorithm with a temporal smoothing procedure which is able to exploit the strong temporal coherence which is normally present in a video.

We have demonstrated state-of-the-art results on the DUT-WEBV dataset and we have shown an extensive analysis of the contribution given by different web sources. 
We plan to extend this work with a large experimental campaign with an open set of tags (not only the ground truth labels provided in the original benchmark) in order to evaluate our system in a tag recommendation scenario.

\section*{Acknowledgments}
This research was supported in part by a grant from the Tuscany Region, Italy, for the AQUIS-CH project (POR CRO FSE 2007-2013).
L.~Ballan acknowledges the support of a Marie Curie Individual Fellowship from the EU's Seventh Framework programme under grant agreement No.~623930.





\bibliographystyle{model1-num-names}
\bibliography{videotags}





\vspace{-10pt}

\section*{Vitae}

\vspace{-5pt}
\small{\noindent{\bf Lamberto Ballan} is currently a postdoctoral researcher at Stanford University, supported by a prestigious Marie Curie Fellowship by the European Commission. He received the Laurea and Ph.D. degrees in computer engineering in 2006 and 2011, both from the University of Florence, Italy. He was a visiting scholar at Telecom ParisTech in 2010. He received the best paper award at the ACM Workshop on Social Media 2010, and is also the lead organizer of the Web-scale Vision and Social Media workshops at ECCV 2012 and CVPR 2014.}

\balance
\vspace{10pt}
\small{\noindent{\bf Marco Bertini} is an assistant professor in the Department of Information Engineering at the University of Florence, Italy. His research interests include content-based indexing and retrieval of videos and semantic web technologies. He received the Laurea and Ph.D. degrees in electronic engineering from the University of Florence, Italy, in 1999 and 2004, respectively. He has been awarded the best paper award by the ACM Workshop on Social Media in 2010.}

\vspace{10pt}
\small{\noindent{\bf Giuseppe Serra} is an assistant professor at the University of Modena and Reggio Emilia, Italy. He received the Laurea and Ph.D. degrees in computer engineering in 2006 and 2010, both from the University of Florence, Italy. He was also a visiting scholar at Carnegie Mellon University, USA, and at Telecom ParisTech, France. He has published around 40 publications in scientific journals and international conferences. He has been awarded the best paper award by the ACM Workshop on Social Media in 2010.}

\vspace{10pt}
\small{\noindent{\bf Alberto Del Bimbo} is a professor of computer engineering at the University of Florence, Italy, where he is also the director of the Media Integration and Communication Center. He has published more then 300 publications in some of the most distinguished journals and conferences, and is the author of the monograph Visual Information Retrieval. He is an IAPR fellow and an associate editor of Multimedia Tools and Applications, Pattern Analysis and Applications, and International Journal of Image and Video Processing. He was general co-chair of ACM Multimedia 2010 and ECCV 2012.}

\end{document}